\documentclass[12pt]{iopart}

\usepackage{xcolor}
\usepackage{graphicx}
\usepackage{placeins}
\usepackage{lipsum}
\usepackage{amsfonts}
\usepackage[cases]{cases}
\usepackage{graphicx}
\usepackage{epstopdf}
\usepackage{algorithmic}
\usepackage{cleveref}
\usepackage{booktabs}
\usepackage[tableposition=top]{caption}
\crefname{figure}{Figure}{Figure}
\crefname{table}{Table}{Table}
\ifpdf
  \DeclareGraphicsExtensions{.eps,.pdf,.png,.jpg}
\else
  \DeclareGraphicsExtensions{.eps}
\fi

\crefname{hypothesis}{Hypothesis}{Hypotheses}
\usepackage{amsopn}

\makeatletter
\newcommand*{\addFileDependency}[1]{% argument=file name and extension
  \typeout{(#1)}% latexmk will find this if $recorder=0 (however, in that case, it will ignore #1 if it is a .aux or .pdf file etc and it exists! if it doesn't exist, it will appear in the list of dependents regardless)
  \@addtofilelist{#1}% if you want it to appear in \listfiles, not really necessary and latexmk doesn't use this
  \IfFileExists{#1}{}{\typeout{No file #1.}}% latexmk will find this message if #1 doesn't exist (yet)
}
\makeatother

\usepackage{siunitx}

\DeclareSIUnit\days{days}

\def\mlmodel{\Theta}

\def\uicon{\mathbf{u}}

\def\uiconcorr{\bar{\mathbf u}}

\begin{document}

\graphicspath{{figures/}}

\title[Dynamic Deep Learning Based Super-Resolution]{Dynamic Deep Learning Based Super-Resolution For The Shallow Water Equations}

\author{Maximilian Witte$^1$, 
Fabrício R. Lapolli$^2$,
Philip Freese$^3$,
Sebastian Götschel$^3$,
Daniel Ruprecht$^3$,
Peter Korn$^2$,
and Christopher Kadow$^1$}

\address{$^1$ Division data analysis, German Climate Computing Centre (DKRZ), 20146 Hamburg, Germany}
\ead{witte@dkrz.de, kadow@dkrz.de}

\address{$^2$ Climate Variability, Max-Planck Institute for Meteorology, 20146 Hamburg, Germany}
\ead{fabricio.lapolli@mpimet.mpg.de, peter.korn@mpimet.mpg.de}

\address{$^3$ Chair Computational Mathematics, Institute of Mathematics, Hamburg University of Technology, 21073 Hamburg, Germany}
\ead{philip.freese@tuhh.de, sebastian.goetschel@tuhh.de, ruprecht@tuhh.de}

\begin{abstract}
Correctly capturing the transition to turbulence in a barotropic instability requires fine spatial resolution.
To reduce computational cost, we propose a dynamic super-resolution approach where a transient simulation on a coarse mesh is frequently corrected using a U-net-type neural network.
For the nonlinear shallow water equations, we demonstrate that a simulation with the ICON-O ocean model with a \SI{20}{\kilo\meter} resolution plus dynamic super-resolution trained on a \SI{2.5}{\kilo\meter} resolution achieves discretization errors comparable to a simulation with \SI{10}{\kilo\meter} resolution.
The neural network, originally developed for image-based super-resolution in post-processing, is trained to compute the difference between solutions on both meshes and is used to correct the coarse mesh solution every \SI{12}{\hour}.
We show that the ML-corrected coarse solution correctly maintains a balanced flow and captures the transition to turbulence in line with the higher resolution simulation.
After an \SI{8}{\day} simulation, the $L_2$-error of the corrected run is similar to a simulation run on a finer mesh.
While mass is conserved in the corrected runs, we observe some spurious generation of kinetic energy.
\end{abstract}

\vspace{2pc}

\section{Introduction}\label{sec:intro}
Numerical models of the atmospheric and oceanic circulation greatly aid our understanding of  climate~\cite{Randalletal2019} and are instrumental for numerical weather prediction (NWP). 
While there are fundamental differences regarding scales and driving forces between weather prediction and climate simulations, there are many similarities in the underlying equations and numerical techniques.
Several earth system models, like the Icosahedral Nonhydrostatic Model (ICON), are designed to be used for both.
These models comprise two important components, the dynamical core and sub-grid models, so-called parameterizations. 
While the former is responsible for solving partial differential equations (PDEs) on the discrete grid, the latter is responsible for approximating the effect of processes such as diffusion, mixing, or turbulence that can not be resolved explicitly at a given spatio-temporal resolution~\cite{StaniforthThuburn2012}. 
To improve the accuracy of weather predictions and climate projections, there is a strong push towards ever finer grid resolutions~\cite{NeumannEtAl2019}.
However, finer mesh resolutions incur increasing computational cost and runtimes of climate models and energy consumption of high-performance computing (HPC) machines.

Machine Learning (ML) has been gaining popularity 
in many areas of simulation based science. 
ML techniques are applied in NWP and climate sciences in different ways.
Pure data-driven ML approaches, for example, are trained by reanalysis data that contain the optimal combination of model and observational information. 
These systems have started to challenge short and medium term weather predictions in terms of accuracy while being much more efficient~\cite{lam2022,pathak2022}. 
Physics-informed Machine-Learning aims to include physical relationships such as dynamical equations as a constraint within the optimization process~\cite{Kashinath2021}.
ML techniques have also been applied to emulate unresolved physical processes such as convection, closures for eddy parametrizations of the ocean, or replacing specific subcomponents of a climate model \cite{yuval2020,Zhang2023}.

In recent years, the advent of super-resolution has significantly advanced the frontiers of image and video processing. 
Super-resolution attempts to enhance the resolution of an image or video, aspiring to construct a high-resolution output from one or more low-resolution inputs~\cite{Wang2021}. 
It is used across a spectrum of applications, ranging from medical imaging and security surveillance to entertainment~\cite{Khoo2020}.
Central to super-resolution in machine learning is the deployment of deep convolutional neural networks~\cite{Dong2014} (CNNs), Generative Adversarial Networks~\cite{Ledig2017} (GANs) or Diffusion models~\cite{Gao2023}. 
These networks, through training on datasets comprising low-resolution and high-resolution image pairs, have demonstrated remarkable proficiency in generating high-resolution images from low-resolution inputs. 
A blend of super-resolution and image inpainting techniques has shown potential in enhancing outcomes, illustrating the synergy between these approaches. 
An example of this approach within climate science is the work of Kadow et al. 2020 \cite{Kadow2020}, who apply image inpainting techniques, leveraging deep learning algorithms, to mitigate missing data issues in the HadCRUT4 temperature dataset. 
However, the utility of machine learning extends beyond the realms of image resolution enhancement and data gap mitigation and can help with bias correction in numerical climate models~\cite{Moghim2017} or in other related~\cite{Tao2016} and non-related disciplines~\cite{Xu2022}.

The transfer of information to higher resolutions or smaller scales is referred to as downscaling in climate science and can be divided into two main approaches: statistical and dynamical downscaling~\cite{ekstrom2015appraisal}. Statistical downscaling relies on empirical relationships between large-scale atmospheric variables and local climate features, producing fine-resolution data by linking observed local data to coarser model outputs \cite{Maraun2010}. Dynamical downscaling, on the other hand, uses physically based regional climate models (RCMs) to simulate local climate processes based on boundary conditions from global climate models (GCMs), capturing complex physical interactions on a finer spatial scale \cite{xu2019dynamical}. Super-resolution machine learning models for fluid flow downscaling have been extensively studied and summarized by Fukami et al.~\cite{Fukami2023}. The use of con\-vo\-lu\-tional neural networks in a multi-scale design, such as their downsampled skip-connection/multi-scale model (DSC/MS) and U-nets, has been shown to provide superior results compared to simple convolutional neural networks~\cite{Fukami_2019, Fukami2023}. 
Due to their efficient spatial dimension reduction, U-net architectures have been successfully applied to the reconstruction of turbulent flows, showing accurate results with excellent convergence properties and computational performance~\cite{Esmaeilzadeh2020,pant2021}.

In our work, we integrate high-resolution information obtained by a neural network into a low-resolution configuration of the same model. 
Our approach augments a coarse, mesh-based algorithm by frequent modifications of the solution by the deep neural net, designed to improve the ``effective resolution'', that is, correct the coarse solution towards the restriction of a solution computed on a finer mesh.
Since this happens at runtime in the time-stepping loop, we call this approach ``dynamic super-resolution''.
We show that this hybrid system is capable of maintaining a balanced flow state and transition to geostrophic turbulence in a similar way as the high-resolution model. 
This is different from so-called ``nudging-to-fine'' approaches, that use ML to learn a state-dependent bias correction which is included in the source term of the governing PDE~\cite{BrethertonEtAl2022,ClarkEtAl2022}. 

A related hybrid ML-PDE solver was introduced by Pathak et al.\ for a two-dimensional turbulent flow simulation~\cite{pathak2020}, but was based on the fine scale features of a quasi-steady solution, and thus cannot capture  transient behavior like the onset of instabilities.
Recently, Barthelemy et al. investigated a similar approach using a con\-vo\-lu\-tional super-resolution machine learning model to downscale low-resolution data of a quasi-geostrophic test case~\cite{Barthelemy2022}. In contrast to Barthelemy et al.~we omit the use of an Ensemble Kalman filter to assimilate the output of the machine learning model. This allows us to focus  on the impact of the ML algorithm without interference of the Kalman filter's error covariance matrices.
Our approach shares some similarities with the recent work of Margenberg et al.~\cite{Margenberg2024}.
They use a DNN to provide some fine scale information to a multi-grid solver to provide a solution that is more accurate compared to using the coarse grid alone.
This avoids the need to store and process fine mesh data after training is finished but, of course, does not produce the actual fine mesh solution.
Furthermore, in contrast to their architecture, our neural network is purely convolutional, thus independent of the spatial domain, and can be applied localized as well. 
As we do not apply the neural network in the context of a geometric multigrid method, the network does not need access to residuals or cell geometries.
While Margenberg et al.\ focus on reproducing the statistics of turbulence, we study the subtle interplay between balance and loss-of-balance for a geostrophic jet in a deterministic setting.

\section{Methodology}\label{sec:method}

\begin{figure}[!ht]
\def\svgwidth{\linewidth}
\centering
\small{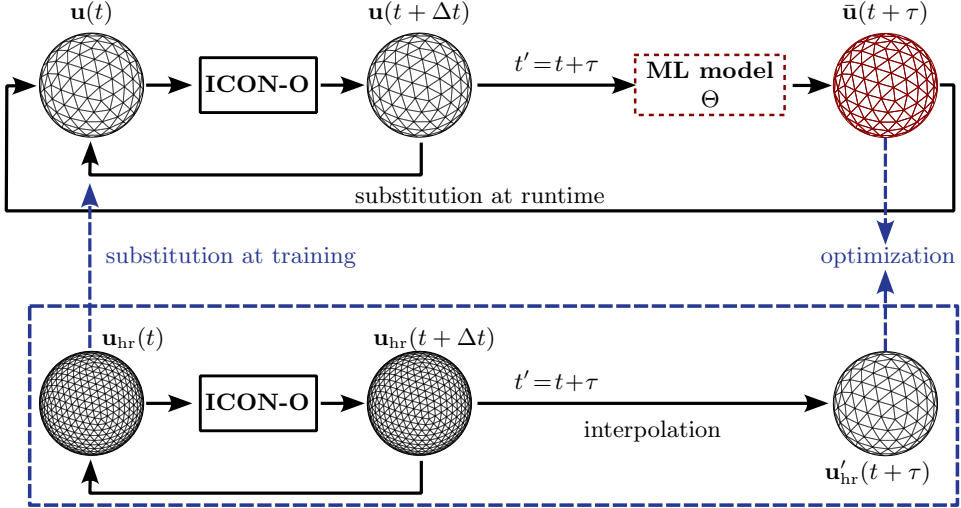}
\caption{Hybrid approach combining numerical simulation using the ICON-O model with machine-learning-based super-resolution.
The data flow during runtime is indicated by the continuous arrows (top panel), whereas the dashed blue lines indicate data movement during training.
$\Delta t$ is the time step used in the numerical simulation (ICON-O).
At $t=0$ and $\tau \gg \Delta t$ we use a distance weighted interpolation method to map the high resolution simulation ground truth onto the low resolution grid for the input and the ground truth of ML-model $\mlmodel$ training.}
\label{fig:approach_w_training}
\end{figure}
The aim of our super-resolution approach is to run a simulation on a coarse mesh while frequently correcting it using a trained ML-model to resemble the restriction of a simulation that was run on a finer mesh. 
This approach does integrate numerical subgrid-scale information into the coarse resolution simulation by correcting the whole state vector. 
This differs from  classical parameterization approaches where specific physical processes are modeled.
Our proposed method is visually summarized in \cref{fig:approach_w_training}, which shows the interaction between the Icosahedral Nonhydrostatic Ocean model (ICON-O) and the ML model $\mlmodel$ during simulation runtime and ML model training.

During runtime, we use a trained ML model to periodically correct the velocity field $\uicon$ at time steps $t'=t+\tau$ via
\begin{equation}
\uiconcorr(t+\tau) = \mlmodel(\uicon(t+\tau)).
\end{equation}
The ML model was previously trained according to the scheme shown in the blue box in~\cref{fig:approach_w_training}. 
We define the initial condition on a high resolution grid $\uicon_{\text{hr}}(t)$ and interpolate it to a low resolution grid $\uicon'_{\text{hr}}(t)$. 
Then we integrate both states using the model time step $\Delta t$ until $t'=t+\tau$. 
We use the output velocity field of the low-resolution simulation $\uicon(t+\tau)$ to train the ML model using the high-resolution, interpolated velocity field $\uicon'_{\text{hr}}(t+\tau)$ as ground truth
\begin{equation}
\min_{\theta} \mathcal{L}\left(\mlmodel(\uicon(t+\tau)),\uicon'_{\text{hr}}(t+\tau) \right),
\end{equation}
where $\theta$ are the parameters of the ML model $\mlmodel$ and $\mathcal{L}$ the loss function which is minimized during training.
In the next sections, we elaborate on the details of our method, the ICON-O model, the ML architecture, its training and the experiment to show the potential of our hybrid approach.

\subsection{ICON-O Numerical scheme}
\begin{figure}
    \centering
    \includegraphics[width=.5\linewidth]{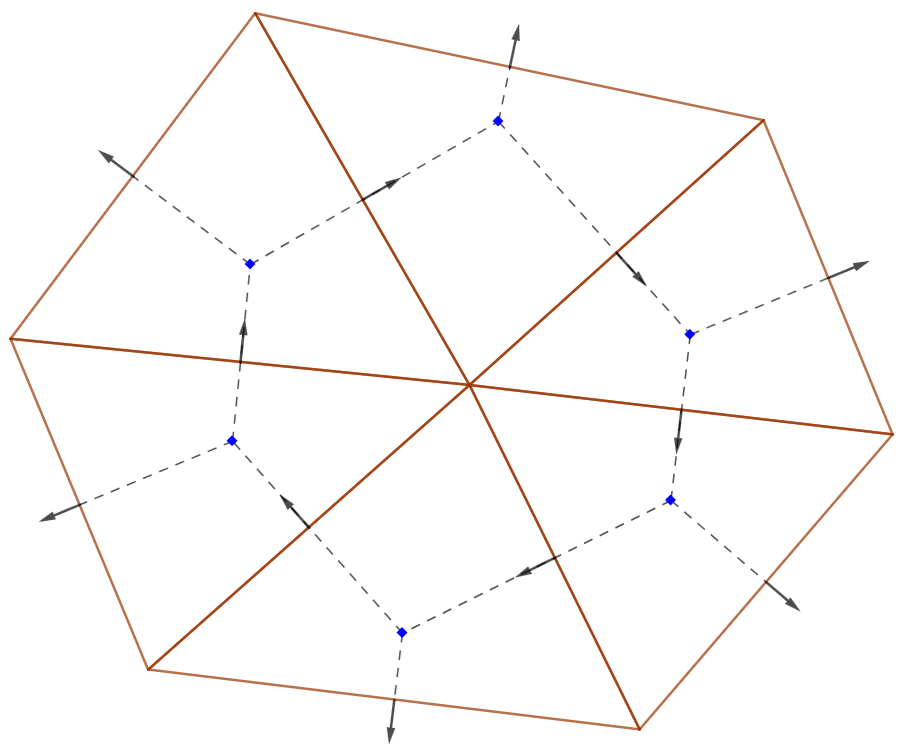}
    \caption{ICON-O's horizontal discretization design. The scalar height field is located in the blue diamond shape, while the vector velocity field is decomposed in the midpoint edges of each triangle.}
    \label{fig:ICONCgrid}
\end{figure}

ICON-O is the oceanic component of the Earth system model ICON~\cite{Hoheneggeretal2022}.
It uses the hydrostatic Boussinesq equations on a sphere with a free surface boundary condition and solves these equations on a triangular grid with Arakawa C-staggering as shown in \cref{fig:ICONCgrid}~\cite{KornLinardakis2018, Kornetal2022}. 

The ocean model equations contain the two-dimensional shallow water equations (SWE) used in this paper as a special case. The SWE are
\numparts 
\begin{eqnarray}               
        \label{equ:shallow_water1} 
        \partial_t h &=& \nabla\cdot \mathbf uh\\
        \label{equ:shallow_water2} 
        \partial_t \mathbf u &=& \nabla [g(h+b)+E_k]+(f+\omega)\mathbf u^\perp,
    \end{eqnarray}
\endnumparts
where $h$ and $\mathbf u = (u,v)$ are scalar height and vector velocity field consisting of zonal ($u$) and  meridional ($v$) velocity components, respectively. Here, $g$ is the Earth's gravity acceleration, $b$ is the topography, $E_k=|\mathbf u|^2/2$ is the kinetic energy, $f$ is the Coriolis parameter, $\omega$ is the relative vorticity, and $\uicon^\perp=\mathbf k\times\mathbf u$ is the perpendicular velocity, where $\mathbf{k}$ is the unitary vertical vector.

The numerical scheme of ICON-O in its shallow-water configuration satisfies discrete conservation laws for mass, total energy, potential vorticity and potential enstrophy~\cite{KornLinardakis2018}. 
A study of its accuracy was provided by Lapolli et al.~\cite{Lapollietal2024}.     

\subsection{Network architecture}
\begin{figure}[th]
\def\svgwidth{\linewidth}
\centering
\small{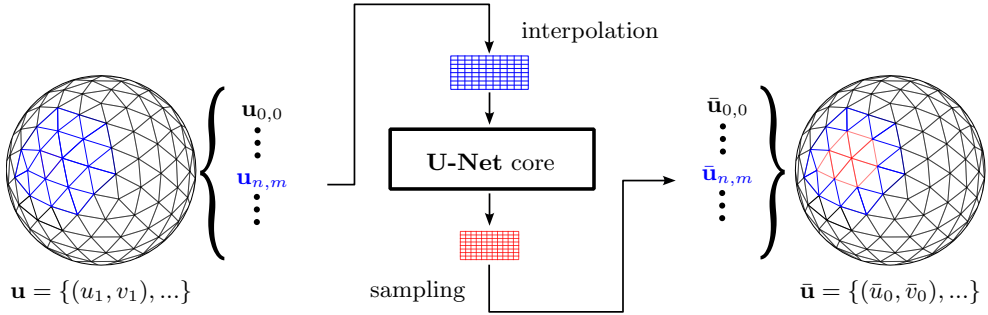}
\caption{Our super-resolution network consists of a U-Net core model and two modules handling the transition between the ICON-O grid and the regular grid. Nearest neighbor interpolation is used to map the local patch $\uicon_{n,m}$ to a regular grid. The corrected patch is obtained by sampling data points from the output of the U-Net according to their coordinates.}
\label{fig:ML_correction}
\end{figure}

We assume that our choice of $\tau$ is sufficiently small so that the difference between the integrated high-resolution and low-resolution velocity fields at a given location depends only on a local neighborhood. We therefore define the ML model on local domains as shown in~\cref{fig:ML_correction}.
The advantage of this approach is that the model otherwise tends to overfit to global structures. 
In addition, model prediction of local patches can be done in parallel, which is particularly useful when the model is running on high performance computing systems.

We define a local patch by $\uicon_{n,m}$ with $n,m \in \{0, \dots,7\}$ as a subset of the unstructured grid of ICON-O. 
We use nearest neighbor interpolation to interpolate the subset to a high-resolution regular grid of $1024\times 1024\,$pixels. 
We use a $10\%$ overlap of the patches during training. 
The output data $\uiconcorr_{n,m}$ is sampled from the output of the core model based on their relative coordinates.

The core model has a U-net structure~\cite{ronneberger2015unet} with skip connections to take advantage of the computational efficiency of convolutions in image to image tasks. 
We define the convolutional layers of the U-net without biases. 
As a consequence, the trivial field correction is itself a trivial field $\Theta(\mathbf{0})=\mathbf{0}$. 

\begin{figure}[h]
\def\svgwidth{\linewidth}
\centering
\small{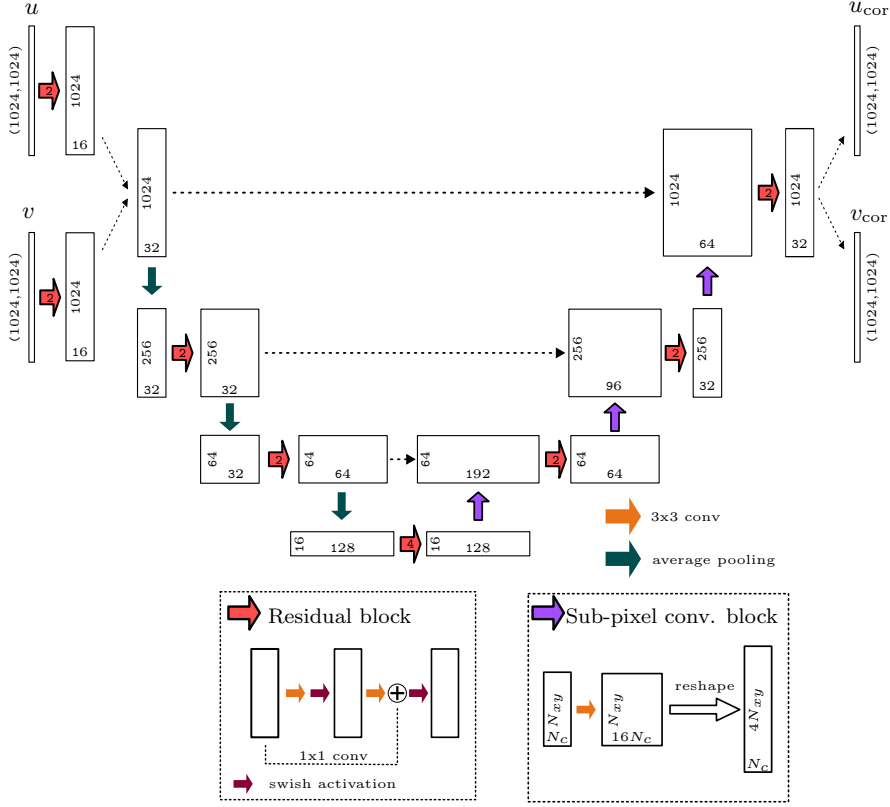}
\caption{U-Net core of the ML model. In the encoder and decoder branch, we use a kernel size of $7\times7$ and $3\times3$, respectively. The dotted arrows indicate concatenation of tensors.}
\label{fig:unet_core}
\end{figure}

The detailed structure of the U-net is shown in~\cref{fig:unet_core}. 
The basic element of our U-Net consists of two consecutive residual blocks, or ResNet blocks~\cite{he2015deep}, with a swish activation function~\cite{ELFWING20183} and a $1\times 1$ convolutional layer for the residual connection. 
We use average pooling layers with a kernel size of $4$ to effectively decrease the spatial dimension. 

Upsampling, i.e., increasing the spatial dimension in the decoder part of the U-Net, is realized by sub-pixel convolutional blocks~\cite{Shi2016}. These consist of a convolutional layer, which increases the channel dimension by a factor of $r^2$, where $r$ is the factor by which the resolution is to be increased in both spatial dimensions. 
A consecutive pixel-shuffle operation rearranges the tensor with a shape of $(r^2 N_c,N_x,N_y)$ such that the channel dimension is decreased by a factor of $1/r^2$, whereas both spatial dimensions are increased by a factor of $r$. We use the kernel weight initialization proposed by Aitken~\cite{Aitken2017} to ensure the absence of checkerboard artifacts.

The output of the encoder branch is concatenated with the respective upsampled output of the decoder layer and fed into the next layer. 
Note that for the input and output, we are using grouped convolutions to efficiently learn input and output features for both velocity components.

\section{Experiment Design}
We analyze the performance of our approach for the Galewsky testcase ~\cite{Galewskyetal2004}.
 This testcase describes the transition from a geostrophically balanced steady state to a barotropic instability. 
 This transition is triggered by a small perturbation that is added to the balanced initial condition and that needs several days to build up.   
The instability of barotropic jets is a classical and well studied process in geophysical fluid dynamics that occurs in both the atmosphere and the ocean. 

The initial condition is defined as follows
\begin{equation}
    u(\phi) = \begin{cases}
        0 & $\phi\leq\phi_0$\\
        u_{\max}/e_n \cdot \exp{\left(1/[(\phi-\phi_0)(\phi-\phi_1)]\right)} & $\phi_0\leq\phi\leq\phi_1$\\
        0 & $\phi\geq\phi_1$,
   \end{cases}
    \label{equ:galewsky}
\end{equation}
where $u_{\max}=80$ ms$^{-1}$, $e_n=\exp\left(-4/(\phi_1-\phi_0)^2\right)$, $\phi_0=\pi/7-n_{\phi}\pi/36$, $n_\phi\in \{-1,0,1\}$, and $\phi_1=\pi/2-\phi_0$. 
This is balanced by the height field
\begin{equation}
    gh(\phi) = gh_0-\int^{\phi} au(\phi')\left[f+\frac{\tan \phi'}{a}u(\phi')\right]d\phi',
\end{equation}
where $a=6.371\times 10^6$ m is the radius of the Earth and the resultant integral is a function of the latitude $\phi$.
Here, the integral is unfeasible to calculate analytically, therefore, we chose to compute it using Rombergs' quadrature, from the bounds of 90$^\circ$S, where $u=0$ to $\phi$.

To trigger the instability and development of turbulence we add the perturbation
\begin{equation}
    h'(\lambda, \phi)= \hat h \exp\left(-[(\lambda_2-\lambda)/\alpha]^2\right)\exp\left(-[(\phi_2-\phi)/\beta]^2\right)\cos\phi,
\label{eq:perturbation}
\end{equation}
where $\lambda$ is the longitude, and $\lambda_2=n_\lambda\pi/4-\pi$, $n_\lambda\in \{0, \dots,7\}$, $\phi_2=\pi/4-n_\phi \pi/36$, $\alpha=1/3$, $\beta=1/15$, and $\hat h = \SI{120}{m}$.

\subsection{Training of the ML model}
\label{section:ML_training}
We employ the training scheme shown in~\cref{fig:approach_w_training} (blue box). 
Starting from the same initial conditions, we integrate the shallow water equations~(\ref{equ:shallow_water1}),~(\ref{equ:shallow_water2}) on both, the high-resolution (\SI{2.5}{\kilo\metre}) and the low-resolution (\SI{20}{\kilo\metre}) grid. 
At $t = \tau=\SI{12}{h}$ we generate outputs from both grids $\uicon_{\text{hr}}(t+\tau)$ and $\uicon_{\text{lr}}(t+\tau)$. 
We obtain the ground truth $\uicon'(t+\tau)$ by interpolating the \SI{2.5}{\kilo\metre} output to the \SI{20}{\kilo\metre} grid using distance weighted interpolation \textit{Climate Data Operators} (CDO)~\cite{schulzweida2019cdo}. We normalise both velocity components by the \SI{95}{\percent} quantile of their magnitudes calculated on the training data distribution. Note that the sign of both components is preserved.
After forward feeding the ML model with the low resolution output $\uiconcorr=\mlmodel(\uicon_{\text{lr}}(t+\tau))$, we calculate gradients based on the loss
\begin{equation}
\mathcal{L} = \mathcal{L}_\text{abs} + \gamma \mathcal{L}_\text{rel}
\label{equ:loss}
\end{equation}
where
\begin{equation}
\mathcal{L}_\text{abs} = \frac{1}{N}\sum_{i}^N\left(|\bar{u}_i - u'_i|\right)^2 + \left(|\bar{v}_i - v'_i| \right)^2
\label{equ:absolute_loss}
\end{equation}
and
\begin{equation}
\mathcal{L}_\text{rel} = \frac{1}{N}\sum_{i}^N \left( \min \left[1, \frac{|\bar{u}_i - u'_i|}{u'_i + \epsilon}\right] + \min \left[1, \frac{|\bar{v}_i - v'_i|}{v'_i + \epsilon}\right]  \right).
\label{equ:relative_loss}
\end{equation}
Here, $u'_i, v'_i$ and $\bar{u}_i,\bar{v}_i$ denote the i-th element of the zonal and the meridional velocity components and $\epsilon=\SI{e-12}{}$ an offset to avoid division by zero.
We chose a squared error loss in the calculation of~\cref{equ:absolute_loss} as an absolute error measure to penalize large valued errors. 
In addition, we add a non-squared relative error loss to \cref{equ:loss} to account for the imbalance between both velocity components. We chose a scaling factor of $\gamma=\SI{0.1}{}$ such that the initial magnitudes of both loss terms are approximately the same. Especially at early stages of the experiment, $v$ is orders of magnitude lower than $u$ but plays a crucial role in disturbing the geostrophic flow. To stabilize the training process, we restrict maximum relative errors to $1$ (\cref{equ:relative_loss}). Note that we use normalized dimensionless quantities in the calculations of \cref{equ:absolute_loss,equ:relative_loss}.

The training data set of the ML model is provided by running variations of the original Galewsky test by modifying the location of the jet~\cref{equ:galewsky} by setting $n_\phi \in \{-1,1\}$, and the location of the perturbation~\cref{eq:perturbation} by using $n_\lambda \in \{0, \dots,7\}$.
For validation, we use the original Galewsky test case with $n_\phi=n_\lambda=0$.
The choice for training and validation data is summarized in \cref{table:training_data}.
We run the simulations over \SI{20}{\days} with a timestep of \SI{10}{\second} and an output frequency of \SI{12}{h}. 
\begin{table}[]
\begin{tabular}{cccc} 
\toprule
  & jet location& perturbation location&  total number of snapshots \\
  & $n_\phi$ & $n_\lambda$&   \\
 \midrule
 training & $ \{-1, 1\}$ & $\{0, \dots,7\}$ & $1184$\\ 
 validation & 0 & 0 & 37\\ \bottomrule
\end{tabular}
\caption{Simulation setups for generating the training and validation data. We use a \SI{12}{h} output frequency and a maximum simulation time of \SI{20}{days}.}
\label{table:training_data}
\end{table}
The total number of unique patches, which are randomly sampled, amounts to $N= \SI{32}{(patches)}\cdot\SI{1184}{(snapshots)}=37888$ and $2368$ for training and validation, respectively.
We train the model for \SI{50000}{} mini-batch iterations with a batch size of $32$ on $4\times \text{A100}$ GPUs using a decaying learning rate with an initial learning rate of $10^{-4}$ (\cref{fig:learning_progress_ml}). 

\subsection{ML correction during runtime}
Given the trained ML model, we are now running ICON-O coupled with the ML model (\cref{fig:approach_w_training}, top panel only). We again use the original Galewsky test case with $n_\lambda=n_\phi=0$, a model time step of \SI{10}{s} and a correction frequency of \SI{12}{h}.

For evaluating the performance of the ML coupling with ICON, we compute the accuracy error norms for the simulated fields, i.e., velocity and height. 
We compute errors in both $L_2$-~and $L_{\text{max}}$-norm, as they will provide information about the integrated error as well as local errors.
These norms are defined as:
\begin{subequations}
    \begin{equation}
        L_2=\sqrt{\frac{S[(f^{[n]}-f^{[r]})^2]}{S[(f^{[r]})^2]}},
    \end{equation}
    \begin{equation}
        L_{\max}=\frac{\max_j \left|f^{[n]}_j-f^{[r]}_j\right|}{\max_j\left|f^{[r]}_j\right|},
    \end{equation}
\end{subequations}
where $f^{[n]}_j$ and $f^{[r]}_j$ is the reference (high resolution) and coarse fields at cell $j$, respectively. $S$ is defined as
\begin{equation*}
    S[f] = \frac{\sum_j A_jf_j}{\sum_j A_j}
\end{equation*}
where $A_j$ is the area of the $j$-th cell. 
The simulated field is the low resolution (LR \SI{20}{\kilo\metre}) simulated outputs, while the reference solution is the interpolated high resolution (HR \SI{2.5}{\kilo\metre}) to LR simulated outputs. 
For a proper comparison, we interpolate the HR outputs with a distance weighted interpolation from the \textit{Climate Data Operator} (CDO) software.

Additionally, to evaluate the conservation properties of our schemes we track the energy over time.
The calculations of the kinetic and potential energy are done by the following formulae
\begin{eqnarray*} 
    \overline{E_k} = \sum_j \frac{|\mathbf u_j|^2}{2}A_j, \quad\text{and}\quad 
    \overline{E_p} &=& g\sum_j h_jA_j.
\end{eqnarray*}

Another important metric to evaluate is the transfer of energy between different scales within the simulation, which for the 2D fluid simulation follows well known scaling laws (see, e.g., \cite{Kraichnan1967}).
We evaluate this metric analyzing the spectra of kinetic energy and enstrophy.
This is done by decomposing the velocity in its divergence and vorticity components, and, for each component expanding them in their respective spherical harmonic components
\begin{equation}
    \psi=\sum_m\sum_n\psi_n^m P_n^m(\phi)e^{im\lambda},
\end{equation}
where $\psi$ is either divergence or curl of the velocity, $\psi_n^m$ are the coefficients, and $P_n^m$ are the associated Legendre functions.
The spectra are given by the coefficients $\psi_n^m$, which can be obtained, analogously to the Fourier transform, as
\begin{equation}
    \label{eq:spectracoeff}
    \psi^m_n=\int^1_{-1}\int_0^{2\pi}\psi e^{-im\lambda}\ d\lambda P^m_n(\phi) \ d\phi.
\end{equation}
{In order to calculate the coefficients, we use nearest neighbor interpolation  to a Gaussian Grid \cite{HortalSimmons1991} {and Fourier transform in the zonal component. We then compute the meridional component by exploiting the zeroes of the Legendre polynomial using the Gaussian quadrature method.}}
{Finally, the spectra are obtained by the following equations}
{\begin{equation}
    \text{KE}_n=\frac{a^2}{n(n+1)}\left[|\zeta_n^0|^2 + \sum_m|\zeta_n^m|^2 + |\delta_n^0|^2 + \sum_m|\delta_n^m|^2\right],
\end{equation}}
{\begin{equation}
    \text{EN}_n=|\zeta_n^0|^2 + \sum_m|\zeta_n^m|^2,
\end{equation}}
{where $\text{KE}_n$ and $\text{EN}_n$ are the kinetic energy and enstrophy spectra}{; $\zeta_n^m$ and $\delta_n^m$ are the spectra coefficients of the vorticity and divergence obtained from \cref{eq:spectracoeff} and $a$ is the radius of the earth.}

\section{Numerical Accuracy}\label{sec:results}
We first analyze the performance of the ML model correcting the low resolution Galewsky test case that periodically has been initialized by a high resolution run as shown in~\cref{fig:approach_w_training}. This serves as a best case for the ML correction performance, as the re-initialization from high resolution avoids error propagation.
We then compare these results to the coupled run, where the output of the ML is used for initialization~\cref{fig:approach_w_training} (top panel only).

\subsection{ML Training}
\begin{figure}[th]
    \centering
    \small{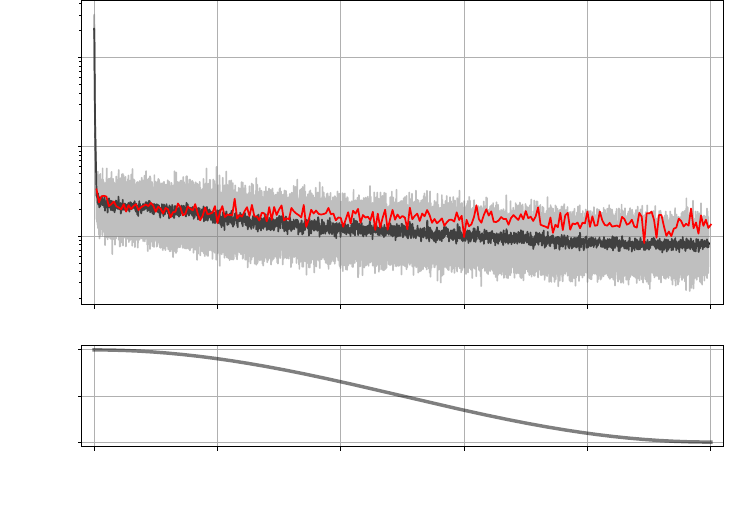}
    \caption{ML model learning progress as a function of mini-batch iteration. Validation was done at a frequency of \SI{200}{} iterations, averaged over \SI{10} iterations. To compare both losses, we applied a moving average window to the training loss. The learning rate as a function of the training iteration is shown in the bottom graph.}
    \label{fig:learning_progress_ml}
\end{figure}

The training progress is shown in \cref{fig:learning_progress_ml}. 
The presence of large fluctuations in the training loss is because of the random sampling of local patches. 
Even though we restrict maximum losses to values of 1, low amplitudes of $v$ result in a high contribution to the relative error~\cref{equ:relative_loss}. 
To avoid overfitting of the data, we stopped the optimization after \SI{50000}{} mini-batch iterations.

\begin{figure}[th]
    \centering
    \includegraphics[width=\linewidth]{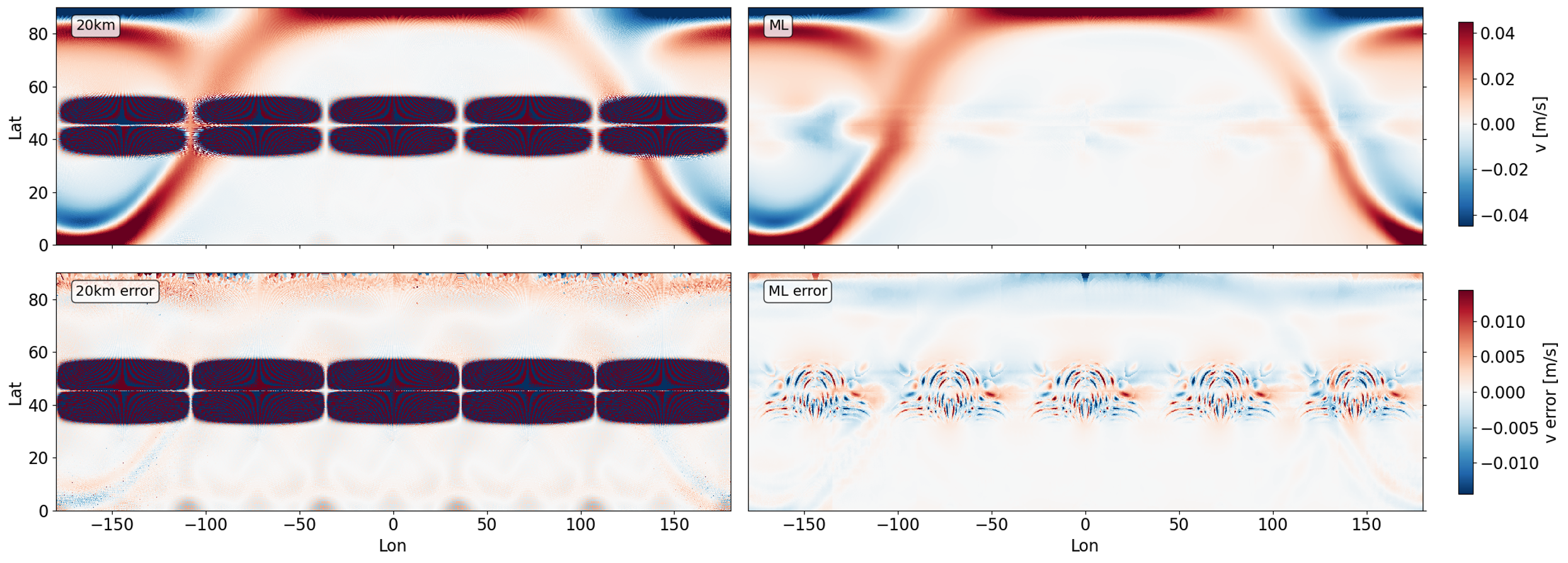}
    \caption{$v$-component of the velocity field at $t=12$h (\textit{top row}) and corresponding error maps (\textit{bottom row}) of the \SI{20}{\kilo\metre} simulation (\textit{left}) and the ML output (\textit{right}). The integration of the same initial condition ($t=0$) on a \SI{2.5}{\kilo\metre} grid serves as ground truth. Refer to \cref{fig:approach_w_training} for further details of the training scheme. {The centered patterns in the left panels indicate numerical oscillations induced by integration on a coarse resolution grid (\SI{20}{\kilo\metre}).}}
    \label{fig:ml_error_12h}
\end{figure}

\begin{figure}
    \centering
    \includegraphics[width=\linewidth]{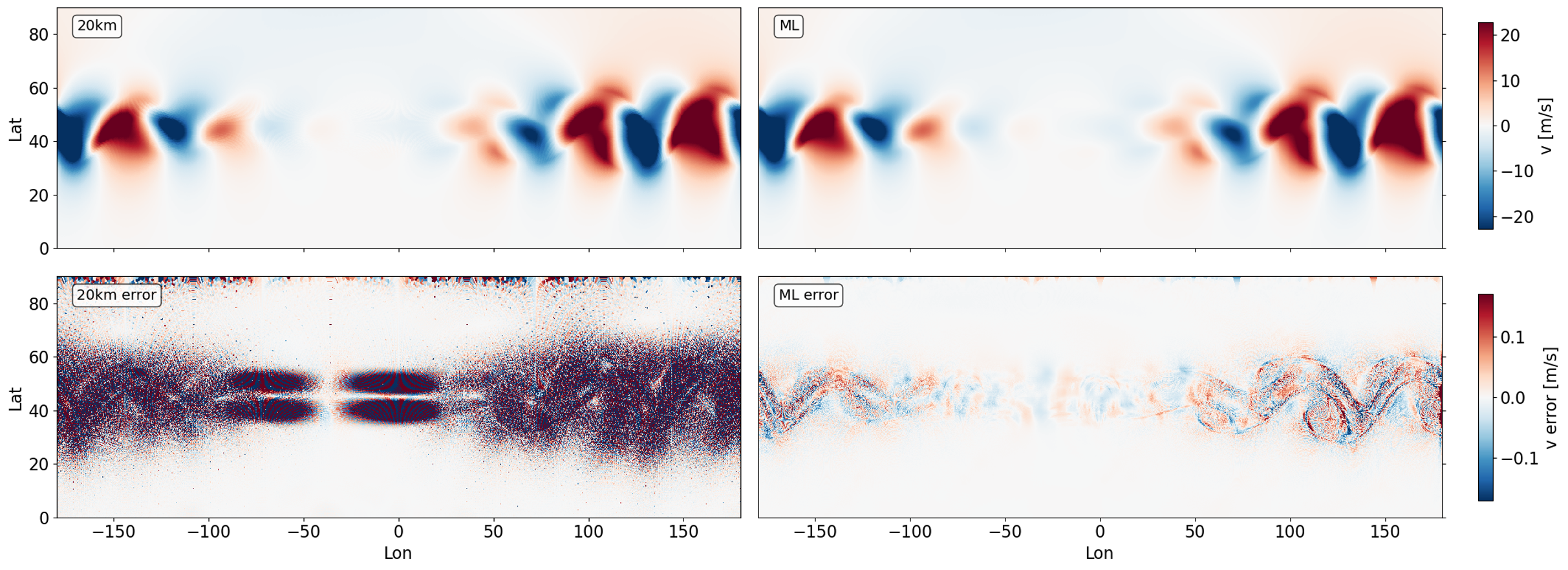}
    \caption{$v$-component of the velocity field at $t=7$d (\textit{top row}) and corresponding error maps (\textit{bottom row}) of the \SI{20}{\kilo\metre} simulation (\textit{left}) and the ML output (\textit{right}). The integration of the same initial condition ($t=\SI{6.5}{\days}$) on a \SI{2.5}{\kilo\metre} grid serves as ground truth. Refer to \cref{fig:approach_w_training} for further details of the training scheme.}
    \label{fig:ml_error_7d}
\end{figure}

\Cref{fig:ml_error_12h,fig:ml_error_7d} show the global snapshots of the uncorrected ML input and the ML output of the $v$-component of the velocity field $\uicon$ at $t=\SI{12}{h}$ and $t=\SI{7}{\days}$. 
The ML model is able to remove the numerical oscillations induced by the coarse resolution at both time points, with $v$ ranging over several orders of magnitude. 
{At early time steps, these numerical oscillations are particularly pronounced around the centre of the jet, where large, high-frequency patterns appear in the left panels of \Cref{fig:ml_error_12h}. The global output of the ML model is smooth and shows no pattern to indicate local patches. Thus, our ML model is robust to small variations of the input data in neighbouring patches.}

\subsection{Numerical Accuracy}

\begin{figure}
    \centering
    \includegraphics[width=\linewidth]{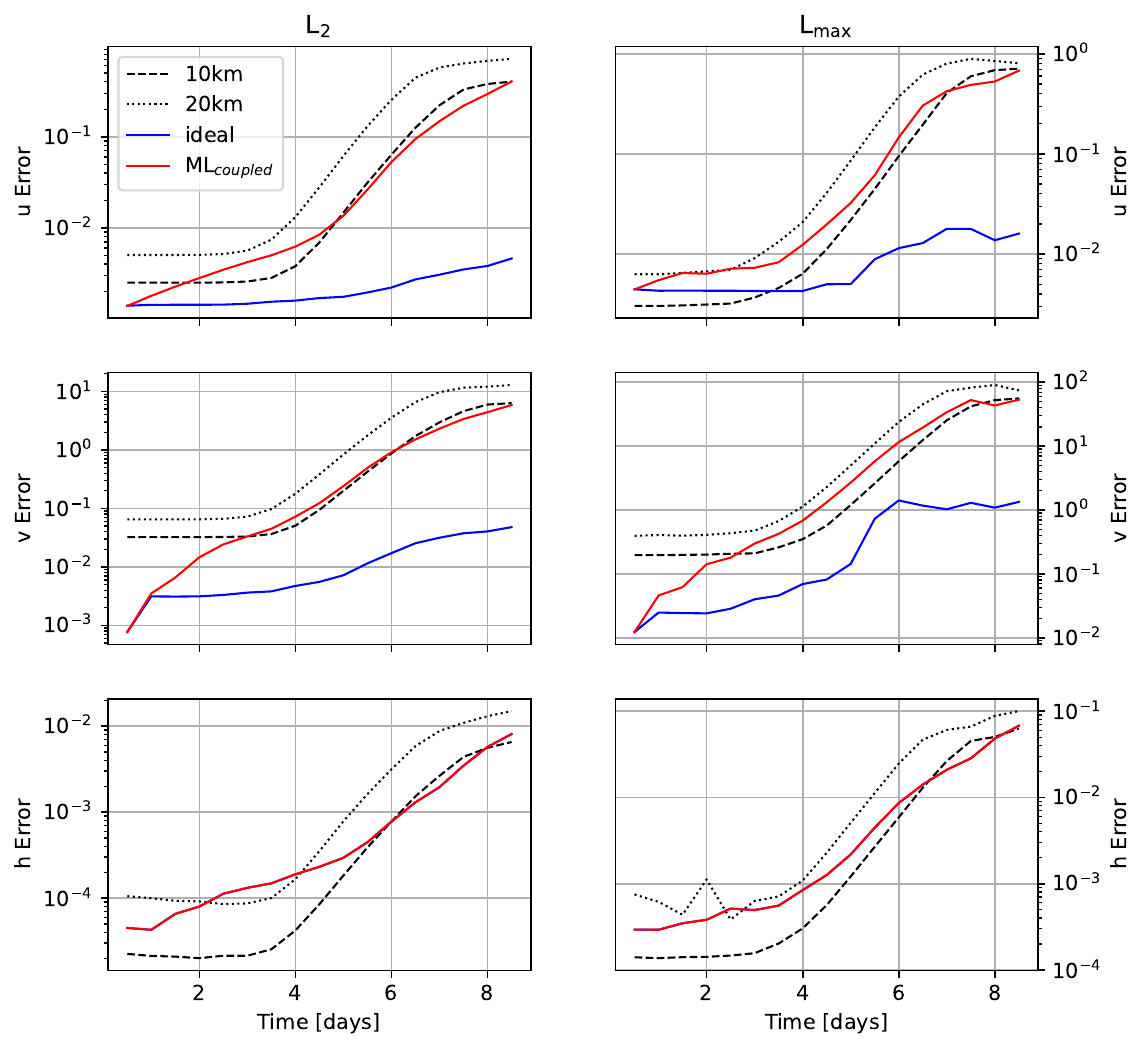}
    \caption{Error for the ICON-O simulations with \SI{10}{\kilo\metre} and \SI{20}{\kilo\metre} resolution, the ICON-O simulation with ML replacement every \SI{12}{h} (ML$_{coupled}$). The \textit{ideal} curve shows the error of the ML model on the validation set, which results from correcting the \SI{12}{h} output of the \SI{20}{\kilo\metre} that is periodically initialized by the high-resolution \SI{2.5}{\kilo\metre} ground truth. See \cref{section:ML_training} for further details. Note that the blue curve is missing in the sea surface height error h, since the ML model corrects only the velocity components.}
    \label{fig:accuracyGalewsky}
\end{figure}

The uncoupled ICON simulation shows a small error in both norms for all fields, quickly increasing from day 4 to day 8, where it then saturates until the end of the computational simulation, see~\cref{fig:accuracyGalewsky}. 
For both $L_2$ and $L_{\max}$ norms, the errors of the uncoupled simulations for the last day of integration are found in \cref{tab:HRLRerror}.
Note that, since the meridional component of the velocity is zero at the initial state, the $v$-component is not normalized as the other fields. 
Regardless, we observe a near first order accuracy of these fields on both norms, which is expected from the model. 
Additionally, it can be observed that the order of accuracy is slightly lower for the maximum norm.
It may be the case that the testcase we are evaluating is prone to numerical oscillations or that the reference simulation of \SI{2.5}{\kilo\metre} is not fine enough to achieve first order accuracy in the maximum norm.

\begin{table}[]
    \centering
    \begin{tabular}{l  c c c  c c c}
        \toprule
          & \multicolumn{3}{c}{$L_2$}  &  \multicolumn{3}{c}{$L_{\max}$}\\
          \cmidrule(r){2-4}\cmidrule(r){5-7}
          & \SI{10}{\kilo\metre}& ML$_{\text{coupled}}$& \SI{20}{\kilo\metre} & \SI{10}{\kilo\metre}& ML$_{\text{coupled}}$ & \SI{20}{\kilo\metre}\\
         \midrule
         $u$ & 4.40$\times$10$^{-1}$& 4.43$\times$10$^{-1}$ &7.20$\times$10$^{-1}$ & 0.78& 0.76 &1.25 \\
         $v$ & 7.47& 7.58 &13.99 &  58.79& 73.19 &76.59\\
         $h$ & {0.81$\times$10$^{-2}$} & {0.84$\times$10$^{-2}$}& {1.01$\times$10$^{-2}$} & {6.78$\times$10$^{-2}$} & {7.10$\times$10$^{-2}$} & {9.23$\times$10$^{-2}$}\\
         \bottomrule
    \end{tabular}
    \caption{Errors of the uncoupled simulation between \SI{2.5}{\kilo\metre} and \SI{10}{\kilo\metre}/\SI{20}{\kilo\metre} resolutions for the different fields at the last day of integration.}
    \label{tab:HRLRerror}
\end{table}

In contrast, the coupled ICON/ML run has its first correction, displaying a significantly improved error in comparison to the uncoupled \SI{20}{\kilo\metre}. 
The error of the velocity components, although displaying substantially lower values than the uncoupled \SI{10}{\kilo\metre}, quickly increases as the field is integrated in time.
This increase is also reflected in the height field, showing an increase in the error slightly larger than the uncoupled \SI{20}{\kilo\meter} simulation.
{Despite this, as the model is integrated in time, the overall fields' accuracies tend to the order of the uncoupled \SI{10}{\kilo\metre} simulation}.
It is also observed that for some days in the later stages of the simulation, the accuracy of the coupled run overtakes even the uncoupled \SI{10}{\kilo\metre}, especially in the $L_2$ norm of the $u$ and $v$ velocity components.
Additionally, despite the height field not being directly affected by the ML, it is impacted by the corrections performed in the velocity field. 
{The observed deterioration of the error in the early stages of the simulation is driven by the velocity field corrections.
However, as the velocity accuracy improves over time, it adjusts the height field to a lower error, improving its accuracy}.

\begin{figure}[th]
    \centering
    \includegraphics[width=\linewidth]{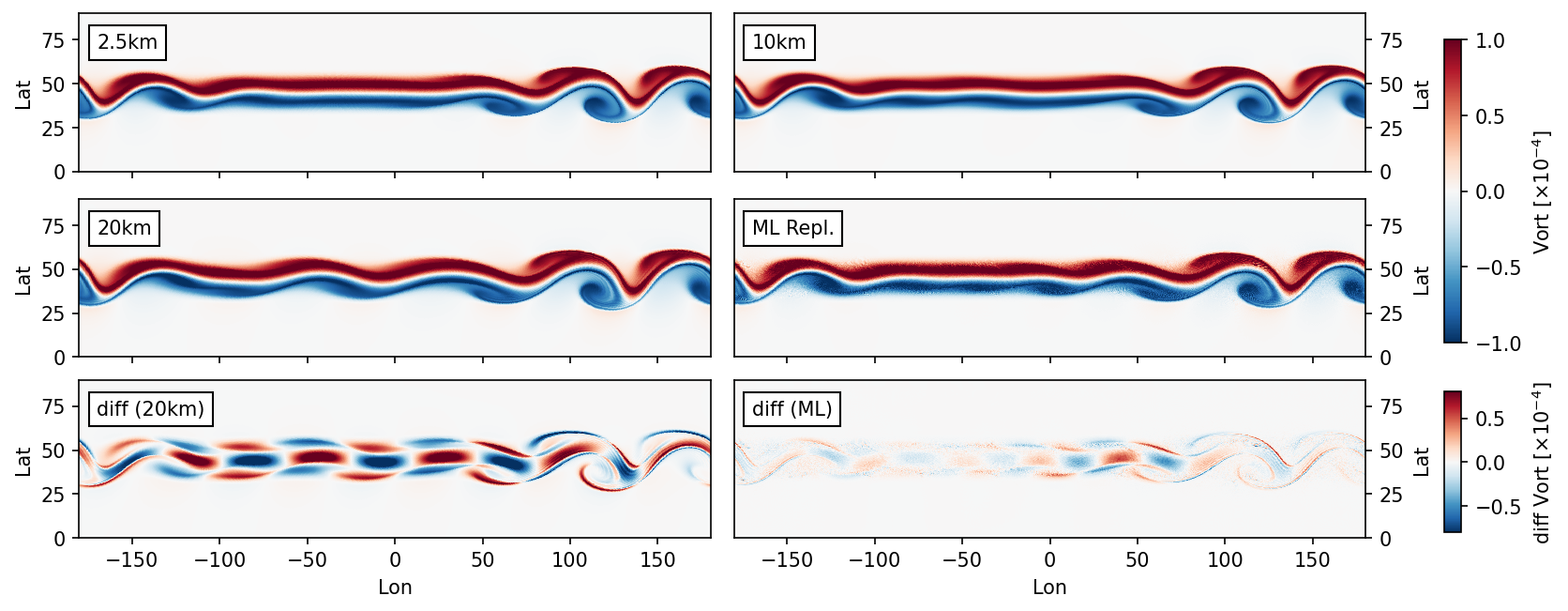}
    \caption{Vorticity map at day 7 of integration for the uncoupled \SI{2.5}{\kilo\metre}, \SI{10}{\kilo\metre}, \SI{20}{\kilo\metre} simulations and the coupled \SI{20}{\kilo\metre} simulation, and the vorticity difference between the uncoupled \SI{20}{\kilo\metre} and \SI{2.5}{\kilo\metre} simulations, and the coupled \SI{20}{\kilo\metre} and uncoupled \SI{2.5}{\kilo\metre} simulations.}
    \label{fig:vorticityGalewsky}
\end{figure}

The improvement in the domain  becomes particularly evident when examining the vorticity field in the 7th day of simulation. 
Here, we observe the early stages of the triggering of the fluid instability.
We see the vorticity field generating vortexes as a consequence of the instability, with these vortexes being transported eastward (see \cref{fig:vorticityGalewsky}).
These vortical formations generate tight gradients which, besides ICON's own numerical noise, are also a source of spurious oscillations \cite{Galewskyetal2004}.
For coarser simulations, these oscillations are greater and the instability is shown to be more developed away from the initial region of perturbation, visible in the uncoupled \SI{20}{\kilo\metre} simulation.

This illustrates the effects of the grid resolution in the fluid simulation. 
From this perspective, it is noticeable that the coupled model accurately portrays the onset of instability and delays local instabilities generated by the interaction between the spurious oscillations of both the unstructured nature of the grid and tight gradients by the testcase.

\begin{figure}[th]
    \centering
    \includegraphics[width=.7\linewidth]{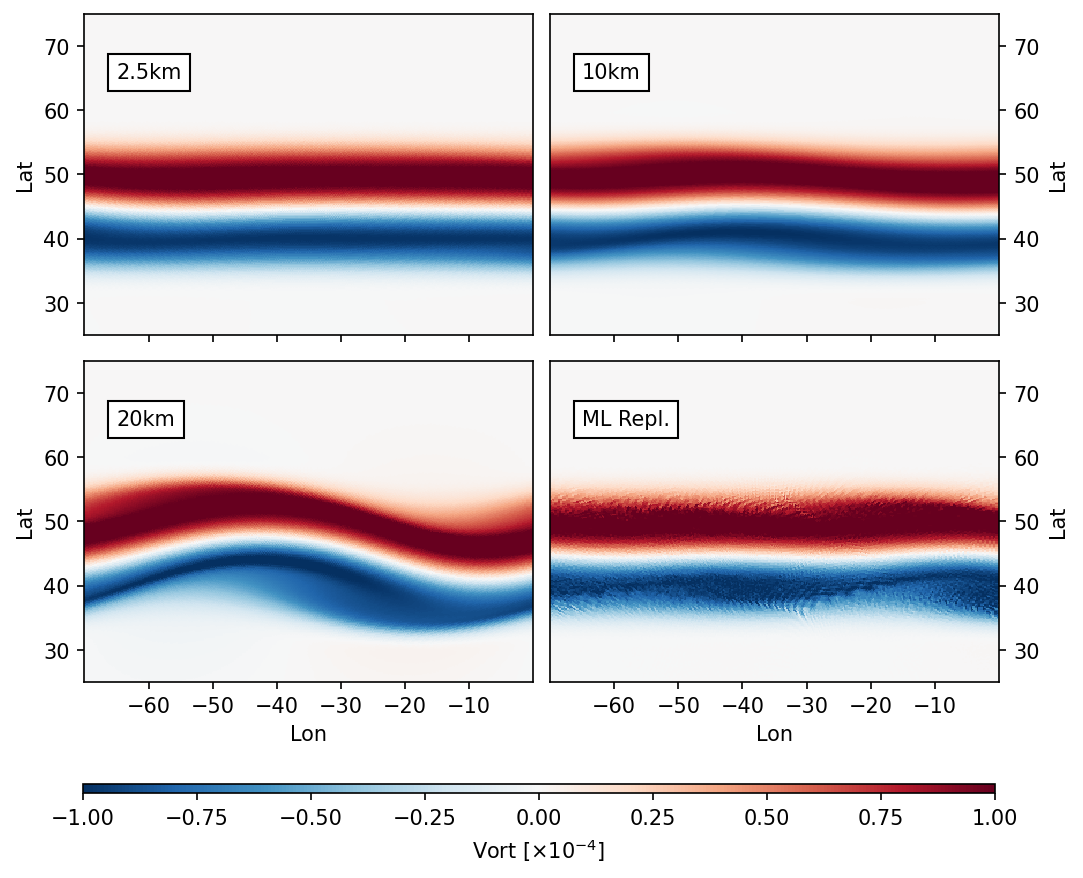}
    \caption{Vorticity as in \cref{fig:vorticityGalewsky}; zoomed at the region 100$^\circ$E-170$^\circ$E and 25$^\circ$N-75$^\circ$N.}
    \label{fig:GalewskyVorticityZoom}
\end{figure}

However, the ML correction add its own set of artifacts, see \cref{fig:GalewskyVorticityZoom}, which is likely playing a role in the accuracy of the model.
Therefore, we may interpret that the difficulty to maintain low error in the first few days is a consequence of these artifacts, especially in the early stages of the simulation.
{As a consequence, due to the lack of diffusivity in the model, these noises do not disappear, and they are still present in the solution and integrated in time.
These oscillations, although being visible, do not amplify nor seem to meaningfully disrupt the solution of the integrated fields, since we can achieve a \SI{10}{\kilo\metre}-order-error simulation}.


\subsection{Energy and Enstrophy}
Despite the improvement, it is important to understand if this artificial impact of the coupled model, especially concerning its numerical oscillations, may disrupt the {energy pathway of the system.
For this purpose we consider the energy and enstrophy spectra. 
From the seminal work by Kraichnan~\cite{Kraichnan1967} the fluid can contain two main inertial ranges with different slopes in the shallow water regime: inverse energy and forward enstrophy cascades with a dissipation rate of $-5/3$ and $-3$, respectively.
We investigate these spectra although they apply to developed turbulence only while the barotropic instability covers the transition to turbulence, because a potential artificial injection of energy or enstrophy due to the coupled model should show up in the spectra.
In this case the spectra would also reveal if this injection occurs at small or large scales.
Due to the local nature of the ML correction we expect changes to appear in the small scales.
For a qualitative assessment whether ICON-O reproduces the theoretical benchmark of the energy and enstrophy spectra the experiment is not appropriate. 
The results of the Galewsky experiment at day 20 display the $-3$ slope through most of the wavenumber space, while the $-5/3$ slope is likely restricted to the low wavenumbers only \cite{Lee2024}.}

\begin{figure}
    \centering
    \includegraphics[width=.7\linewidth]{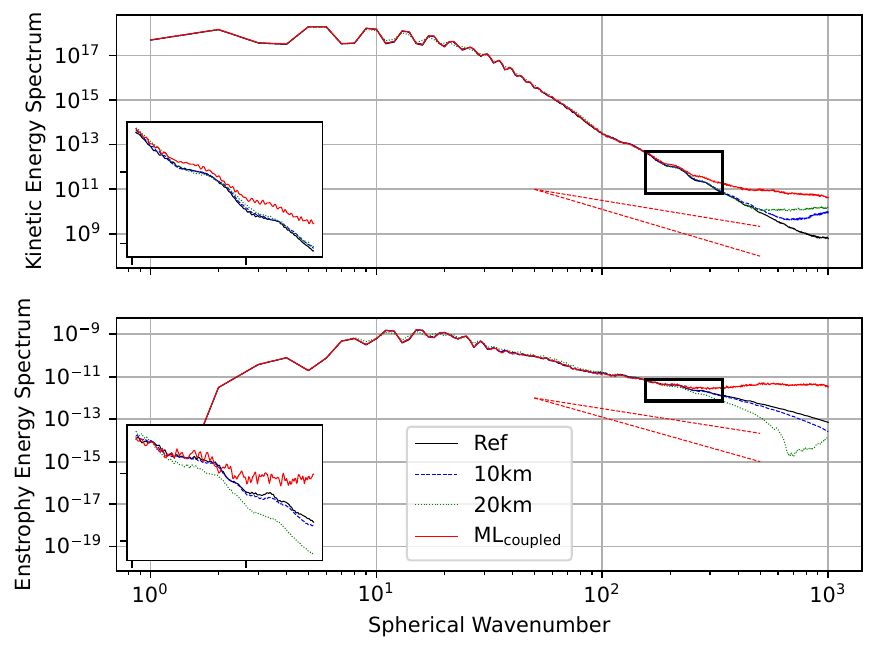}
    \caption{Kinetic energy (top panel) and enstrophy (bottom panel) spectrum of the uncoupled \SI{20}{\kilo\metre} and \SI{10}{\kilo\metre} runs, and the ML coupling simulation. The dashed red lines are the k$^{-3}$ and  k$^{-5/3}$ slopes.}
    \label{fig:spectrum}
\end{figure}

{In our experiments, in day 7, prior to the full turbulence development, we observe a spectrum for the energy of a slope steeper than $k^{-3}$ (\cref{fig:spectrum} upper panel).}

In our analysis, {the slope of the spectra is well represented in all models down to the 300 km wavelength.}
{Towards finer scales, the coupled ML run starts to deviate, displaying an increase in energy, while the uncoupled \SI{20}{\kilo\metre} simulation still maintains the slope along with reference solution.
The deviation of the uncoupled \SI{20}{\kilo\metre} simulation is only detected at index approximately 80 km wavelength.}
This reinforces the argument that the machine learning model is predominantly acting in the smaller scales.

Similarly, the enstrophy spectrum, see \cref{fig:spectrum}, (lower panel), displays a similar enstrophy cascade up to the 200 km resolution.
Around these scales, the uncoupled \SI{20}{\kilo\metre} simulation is the first to deviate from the theory, followed by the coupled \SI{20}{\kilo\metre} simulation.
In these scales, the uncoupled model loses substantially its enstrophy in higher wavenumbers, while the coupled simulation display an increase for higher resolutions.
This indicates that the ML is adding energy into the smaller scales, likely in the form of the previously observed spurious oscillations.
Due to the non-dissipative nature of our system, the high wavenumber regime is not dissipated into mixing and, therefore, remains in the domain.
Although in a more complex model, {these oscillations near grid scale induce spurious mixing of tracers in the ocean \cite{Ducousso2017}}, these can be solved by an appropriate choice of low pass filter, such as a higher order dissipative operator to dim these higher wavenumber waves.

\begin{figure}
    \centering
    \includegraphics[width=.7\linewidth]{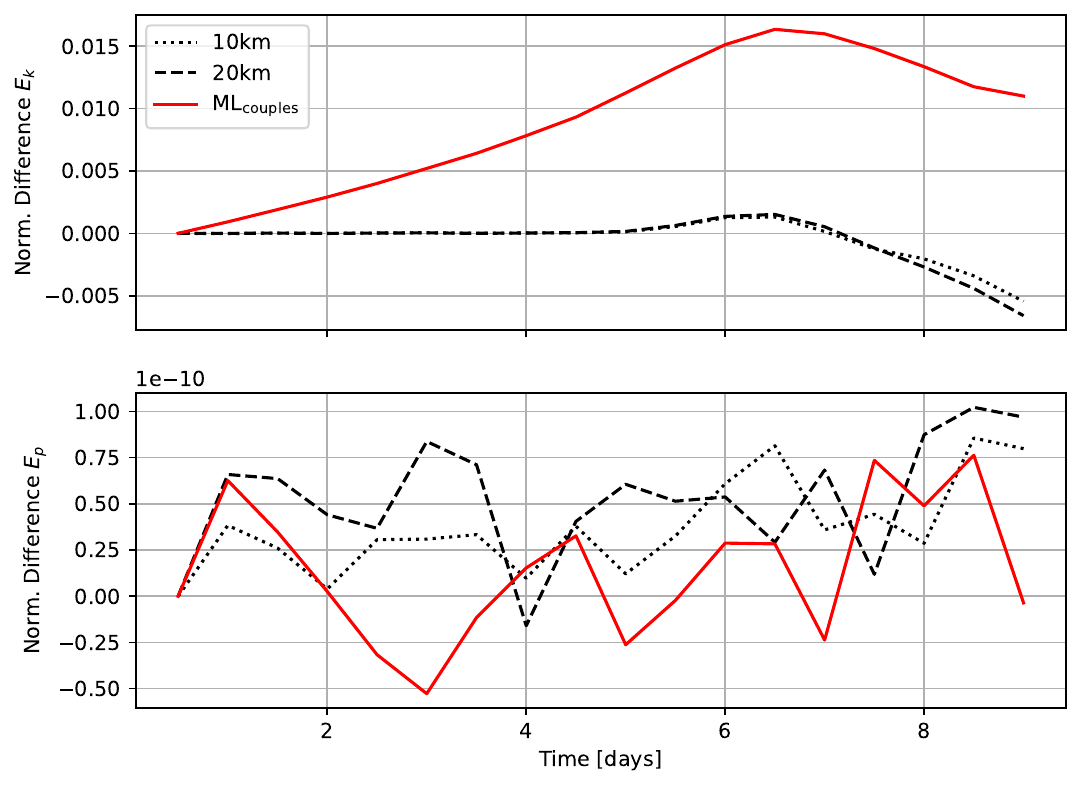}
    \caption{Normalized difference for both globally integrated kinetic (upper panel) and potential energy (lower panel).}
    \label{fig:GalewskyEnergy}
\end{figure}

In order to verify if this increased energy in higher wavenumbers has an overall effect in the energy conservation of our scheme, we evaluate the total energy time evolution (\cref{fig:GalewskyEnergy}).
In the coupled run, it is observed that the kinetic energy of the system increases up to around day 6 and decreases in the subsequent days.
Conversely, both uncoupled simulations show a relatively stable integrated kinetic energy up to day 5, with a slight uptick preceding the instability, followed by a consistent decline in energy, thereafter. 
The integrated energy variation of the uncoupled run is about as half of the coupled runs, indicating that the ML is inputting energy in the domain.
As observed in the energy spectrum, this input of energy is in the form of low resolution (high wavenumbers) waves, likely generated by the spurious oscillations of the ML correction.

\subsection{Runtimes}
{We also measured runtimes of the respective methods for $t=\SI{12}{\hour}$ of integration, subtracting timing for data input and output, for 32 Nodes on the CPU partition of the Levante HPC system (German Climate Computing Center).
While the \SI{10}{\kilo\metre} run took \SI{63}{\second} and the \SI{20}{\kilo\metre} simulation \SI{31}{\second}, the ML-coupled version was done in \SI{55}{\second}.
Hence, we} observe a speedup of \SI{14.5}{\percent} from the ML-coupled simulation compared to the \SI{10}{\kilo\metre} run. 
However, since the focus in this paper was on demonstrating the accuracy of dynamic super-resolution, the ML-corrected model is mostly untuned.
In particular, only CPUs were used since we used a shared partition for the ICON-O simulation and the ML model.
Moving the ML correction to GPUs and optimizing data transfer would require refactoring of ICON-O but could also greatly improve speedup.

\section{Conclusions}\label{sec:conclusions}
We present a hybrid modeling approach, combining a mesh based model with an ML correction to incorporate the net impacts from finer scales, unresolved dynamics.
Because the simulation is time-dependent and the correction is applied after a set number of time steps have been processed, we call the approach dynamic super-resolution.
We demonstrate that the correction allows us to run a global numerical simulation of the shallow water equations on a twice coarser mesh without loss of accuracy. 
The continuity between adjacent predictions in the global output confirms the reliability of employing a local ML model approach to correct a global velocity field.
As our ML model operates solely through convolutions, adapting it  to large-scale fields is straightforward by adjusting the patch size to fit to available computational resources. 
Employing independent patches enables parallelization of ML predictions for efficient corrections. We emphasize that our approach is based on using a single test case and a single set of parameters in terms of resolution difference and temporal range $\tau$. Both the resolution difference and the temporal range will determine the correlation between the low-resolution and the high-resolution sample, which will directly affect the accuracy of the ML model and our general approach.

The trained ML model effectively reduces the discretization errors of the integration of the shallow water equations.
Additionally, using relative errors in optimizing the parameters of the model ensures consistent model performance across different scales, which is crucial to preserve the stability of the jet of the Galewsky test case.
However, the error increases when the output of the ML model is integrated in time in the coupled run. The remaining artifacts produced by the ML model still affect the solution and likely lower the accuracy of the coupling. 
Additionally, the ML model acts as a source of kinetic energy. 
Although it does not push the solution to numerical instability, this could become an issue for more complex regimes. 
Lastly, despite the direct cascade of both energy and enstrophy being respected for large scales, they are affected in the high wavenumbers, likely due to the noise produced by the ML.

In future work, we plan to integrate physical constraints into the ML model to help conserve physical properties in our hybrid approach. 
In addition, training objectives penalizing artifacts and forcing convergence of the ML correction will help to enhance the stability integrating the ML output in time. 
Incorporating statistical ML approaches to estimate model uncertainty may offer further insights into stabilizing the ML correction process. 
Even though we only used CPU resources, we observed a speedup of the coupled simulation compared to a similarly accurate simulation (\SI{10}{\kilo\metre}).
This shows the potential of dynamic super-resolution to reduce wallclock times of the dynamical cores of earth system models while maintaining high accuracy.

\ack
This project received funding from the German Federal Ministry of Education and Research (BMBF) under grant 16ME0679K. Supported by the European Union--NextGenerationEU. This work used resources of the Deutsches Klimarechenzentrum (DKRZ) granted by its Scientific Steering Committee (WLA) under project ID bk1372.

\section*{References}
\bibliography{references}

\end{document}